\documentclass{article}

\usepackage{authblk}
\usepackage{caption}
\usepackage{hyperref}
\usepackage{natbib}

\hypersetup{colorlinks,allcolors=black}

\title{Metadata Might Make Language Models Better}

\author[1]{Kaspar Beelen}
\author[2]{Daniel van Strien}
\affil[1]{The Alan Turing Institute}
\affil[2]{British Library}
\date{Working Paper, November 2022}

\begin{document}

\maketitle







\begin{abstract}
This paper discusses the benefits of including metadata when training language models on historical collections.
Using 19th-century newspapers as a case study, we extend the time-masking approach proposed by \cite{rosin2022time} and compare different strategies for inserting temporal, political and geographical information into a Masked Language Model.
After fine-tuning several DistilBERT on enhanced input data, we provide a systematic evaluation of these models on a set of evaluation tasks: pseudo-perplexity, metadata mask-filling and supervised classification.
We find that showing relevant metadata to a language model has a beneficial impact and may even produce more robust and fairer models.

\end{abstract}


\section{Introduction}

Predictive language modelling has caused a paradigm shift within machine learning and natural language processing.
Especially with the advent of transformer-based architectures, language models increased in size and became more powerful. 
Representations derived from such models are often referred to as ``pixie dust'', which magically improve the performance on almost any NLP application.\footnote{The term ``pixie dust'' was taken from the Stanford ``Natural Language Processing with Deep Learning'' course by C. Manning, https://web.stanford.edu/class/cs224n/}

These revolutions in the field of Deep Learning have trickled down to Digital History.
Slowly but consistently, historians have been interrogating models such as BERT, evaluating their use and assessing their limitations for analysing the textual traces of the past.
Since most off-the-shelf models are trained on contemporary data---often opaque pools of text with little or no information on their provenance---their application to history remains rather problematic.
A steady stream of recent papers started therefore to investigate the merits of ``historicizing'' language models, which entailed training or fine-tuning them on historical corpora. 
These attempts focused on learning representations solely from text, modelling the probabilities of target words only from textual context (either from the surrounding or previous tokens).
However,  extra-linguistic information, often encoded as metadata, can be critical for understanding the meaning of a document. 
This metadata has often been created as the result of significant work by the custodians of these collections and this type of curated metadata is usually not a given for the large web-scraped datasets traditionally used for training language models. 
This paper, therefore, discusses whether models that learn a relationship between textual content and the historical context will yield better representations of language.
Attention to context outside of a text is especially relevant for scholars in the computational humanities and digital history, who strive to situate language use within specific historical, ideological and social configurations.

To study language in relation to certain metadata variables, scholars traditionally slice and dice their data before training models.
To understand diachronic semantic change, \cite{bamler2017dynamic} compared embeddings extracted from models trained on different epochs. 
\cite{azarbonyad2017words} trained models on parliamentary speeches from left- and right-wing MPs to detect ideologically contested concepts.
\cite{montariol2021scalable} proposed incremental fine-tuning as a strategy for detecting language change. 
Instead of splitting the data, the method we present is more holistic: we avoid training separate models for each community or epoch by inserting such relevant contextual information directly into the learning process.
This paper investigates how including such metadata can improve language models, and describes scenarios where this approach can make a difference.
Our approach builds on the recent ``time masking'' technique developed by \cite{rosin2022time}, which proved a simple but effective method for making models such as BERT more sensitive to temporal contexts by pretending year information to the input data.
Our main contributions are threefold:

\begin{itemize}
\item We extend the approach of \cite{rosin2022time} to other extra-linguistic aspects such as political leaning. We refer to our methods as MDMA (MetaData Masking Approaches). 
\item We systematically compare different strategies for metadata masking.
\item We evaluate the benefits of MDMA for different NLP tasks and historical applications.
\end{itemize}


\section{Related Literature}

Whilst a complete overview of developments of language models is out-of-scope for this paper---and would require a much greater word limit---, we provide a summary of key results, particularly as they relate to using language models with historical text. 

The use of language models pre-trained on large datasets has extended to a wide range of NLP tasks. The word2vec \cite{Mikolov2013EfficientEO} and ELMo \cite{Peters2018DeepCW} papers outlined methods that allowed models to be trained on large amounts of language data without supervision through the use of pre-training tasks that only require examples of text data. Taking inspiration from methods developed in the computer vision domain, Howard and Ruder \cite{Howard2018UniversalLM} introduced ULMFIT as a method for enabling effective transfer learning for NLP tasks. Compared to previous approaches such as word2vec, an advantage of this approach is that you would no longer be required to train a model from scratch to target a new domain. This means, for example, a model could more readily be adapted for the historical domain by fine-tuning a `generic' language model trained on Wikipedia. The final major development to note is the mass adoption of transformers-based approaches to language models popularised by `Attention is All you Need'\cite{Vaswani2017AttentionIA}. Partially as a result of the increased computational efficiency of transformer-based language models, the size of these models has grown rapidly with models now passing 500 billion parameters \cite{Smith2022UsingDA}.

\subsection{Language models for historical text}

Following a broader interest in using language models across various downstream tasks, there has been a growing curiosity and exploration of the potential use of language models for working with historical texts. There are several potential challenges to adopting existing pre-trained language models for working with historical texts. Two major challenges are:  

\begin{itemize}
    \item OCR quality: historical text is often made machine-readable through the use of Optical Character Recognition software. The errors introduced by these models may negatively impact the quality of language models trained on this noisy data \cite{vanStrien2020AssessingTI, Jiang2021ImpactOO},
   \item  Semantic change: change in the use of language over time may reduce the performance of models trained on contemporary data when applied to historic texts \cite{Qiu2022HistBERTAP}
\end{itemize}

Various papers have evaluated the suitability of existing language models on historical text data. Since many companies, with resources far exceeding most humanities researchers or libraries, have already trained and released language models, an ideal scenario would be to find that these models can readily be used without the additional expense of training. 

An evaluation of ``zero shot'' approaches, in which contemporary language models are used without additional fine-tuning \cite{Toni2022EntitiesDA}, showed a disappointing performance. An assessment of the effectiveness of using transfer learning for named-entity-recognition found more positive results from levering existing models \cite{Todorov2020TransferLF}. 

Several papers have either fine-tuned an existing language model on historic data \cite{beelen2021time,Hosseini2021NeuralLM} or trained language models from scratch on historic data \cite{Schweter2022hmBERTHM, Bamman2020LatinBA, Aprosio2022BERToldoTH}. An evaluation comparing these approaches \cite{Manjavacas2022AdaptingVP} found that training from scratch led to improved performance of these language models. 
Overall, most of the literature explores the application of language models to historic material changes only the collections used for training. 
The training process itself largely remains untouched. 

\section{Metadata Masking}

Only a few recent papers have systematically experimented with including extra-textual information such as time \cite{rosin2022temporal, dhingra2022time} or geographic information \cite{hofmann2022geographic} in language models. 
Methodologically, these approaches focus on either adapting the input data and masking strategies \cite{rosin2022time, dhingra2022time} or the model architecture \cite{hofmann2022geographic,rosin2022temporal}. 
In this paper, we focus on the former set of approaches, which is transferable to other collections and easily implementable in existing frameworks such as the Hugging Face Transformers Python library \cite{wolf-etal-2020-transformers}.

In this paper, we aim to demonstrate that MDMA anchor documents in their historical context by learning a relationship between metadata and content.
The technique is rooted in masked language modelling, which predicts the probability of a randomly masked token based on its surrounding textual context.
In a sequence of length $n$, the probability of a masked word at position $i$ is $p(w_{i}) = p(w_{i} | w_{1}, ... w_{i-1}, w_{i+1},..., w_{n})$. By adding metadata tokens to the input, for example, a label $l$ indexing the political leaning of the newspaper, the metadata becomes part of the computation of $p(w_{i})$:
\begin{equation}
p(w_{i}) = p(w_{i} | l,  w_{1}, ... w_{i-1}, w_{i+1},..., w_{n})
\end{equation}
In cases where we mask the metadata, the task almost resembles supervised classification, in which we produce a label given a sequence of tokens:

\begin{equation}
p(w_{i}) = p(l |  w_{1}, ..., w_{i-1}, w_{}, w_{i+1},..., w_{n})
\end{equation}

Label $l$ can be any (combination of) metadata field(s). 
The time masking approach \cite{rosin2022time}, which adds information such as the year of publication to the input sequence, has proven to work well, especially for newspapers\cite{dhingra2022time}, a genre of historical data that closely registers the pulse of time by reporting on events that occurred in the recent past. 
With time-maksing  we  encode the year by prepending it to \textit{each} text fragment when training the language model. 
The preprocessed text then looks as follows: 

\begin{quote}
    ``1856 [DATE] These proposals, emanating as they do from the Western powers, either are, or they are not, such as Russia ought to accept.''
    \label{quote:crimeanwar}
\end{quote}

After training the model on such examples, we can gauge how predictions for masked tokens change depending on the indicated year. 
Taking an example from the \textit{Cheltenham Chronicle}\footnote{This snippet was not observed in the training set.} from 1866 ``This effort was due her Majesty's being now in constant communication with Lord Derby'', we mask the personal pronoun and calculate how predictions differ based on the marked year of publication.
In our case, where we used 19th-century newspapers for creating a language model (see below), the predictions will reflect whether a King or Queen was sitting on the British throne, i.e. ``1860 [DATE] This effort was due [MASK] Majesty's...'' returns `her' as the most likely filler. 
When we change the time-stamp to 1810\footnote{i.e. ``1810 [DATE] This effort...''}, the model predicts `his' as the correct pronoun.

Vice versa, models trained with MDMA can guess the date of a textual fragment simply by masking the prepended metadata token. 
Revisiting the above quotation about the Crimean war: if we mask the year token at the start of the sequence (replace ``1856'' with ``[MASK]'') the model predicts ``1854'', ``1855'' and ``1856'' as the model probable fillers.
This allows us to automatically date historical articles.
In this paper, we evaluate both capabilities of the metadata-enhanced language models.



\section{Data Description}

To evaluate different strategies for metadata-masking, we concentrated on a small but openly accessible newspaper corpus which was created as part of a Heritage Made Digital (HMD) newspaper digitisation project.\footnote{https://blogs.bl.uk/thenewsroom/2019/01/heritage-made-digital-the-newspapers.html}.
The HMD newspapers comprise around 2 billion words in total, with the bulk of the articles originating from the (then) liberal paper \textit{The Sun}. 
Geographically, most papers are metropolitan (i.e. based in London), but with the inclusion of \textit{The Northern Daily Times} and \textit{Liverpool Standard}, the historical county of Lancashire is also represented.
Even though HMD is an unbalanced dataset---it massively over-represents a London, liberal perspective---it provides nonetheless diverse enough materials for the purposes of our experiments.
Table \ref{table:hmd_overview} contains a more detailed overview of the corpus.

\begin{table*}
\begin{tabular}{llllr}
&       &        &          &       \\
NLP & Title & Politics & Location &  Tokens    \\
2083 & The Northern Daily Times & NEUTRAL & LIVERPOOL &    14.094.212 \\
2084 & The Northern Daily Times & NEUTRAL & LIVERPOOL &    34.450.366 \\
2085 & The Northern Daily Times & NEUTRAL & LIVERPOOL &    16.166.627 \\
2088 & The Liverpool Standard & CONSERVATIVE & LIVERPOOL &   149.204.800 \\
2090 & The Liverpool Standard & CONSERVATIVE & LIVERPOOL &     6.417.320 \\
2194 & The Sun & LIBERAL & LONDON &  1.155.791.480 \\
2244 & Colored News & NONE & LONDON &       53.634 \\
2642 & The Express & LIBERAL & LONDON &   236.240.555 \\
2644 & National Register & CONSERVATIVE & LONDON &    23.409.733 \\
2645 & The Press & CONSERVATIVE & LONDON &    15.702.276 \\
2646 & The Star & NONE & LONDON &   163.072.742 \\
2647 & The Statesman & RADICAL & LONDON &    61.225.215 \\
\end{tabular}
\caption{\label{table:hmd_overview} Overview of the Heritage Made Digital newspaper collection.}
\end{table*}

\section{Data Processing and Model Training}

While faster than the glacially slow RNNs, training transformer-based language models requires quite some computational resources.
Running a grid search to estimate the impact of every preprocessing strategy and hyperparameter is simply not feasible (unless you have access to a battalion of software engineers and myriad GPUs). 
We want this research to be useful, not wasteful, and, therefore, we were guided by a sense of pragmatism and minimalism in the design of our experiments. 
Many of our decisions to limit the range of our experiments have to be understood in this spirit.
At the same time, we are confident that these limitations don't hurt the quality of our findings.
Drawing on the assumption that ``scaling laws``\cite{Kaplan2020ScalingLF} will mean that relative improvements demonstrated using smaller models and datasets will transfer to larger models, we do not train all the models on the full dataset. 
Moreover, our goal is to evaluate the impact of different preprocessing methods on the performance of language models.
Even though the models themselves remain open to improvement, their \textit{relative performance} will help us to develop best practices. 

In all language modelling experiments, we used the HMD dataset uploaded on the Hugging Face Hub.\footnote{https://huggingface.co/datasets/davanstrien/hmd\_newspapers}
Training a BERT model on the whole corpus would have taken us more than a week (using a reasonably powerful Azure virtual machine with one GPU).
To reduce training time, we made the following decisions:

\begin{itemize}
    \item We downsampled the corpus to half a billion tokens by randomly selecting only 25\% of the articles.\footnote{Data processing was done using the datasets Python library \cite{lhoest-etal-2021-datasets}}
    \item We opted for DistilBERT \cite{Sanh2019DistilBERTAD}, a BERT descendant that contains substantially fewer parameters whilst retaining most of BERT's performance. As a result of the reduced number of parameters, DistilBERT represents a more appropriate choice when training a Masked Language Model with only half a billion tokens.
    \item As we lacked the data to train from scratch, we started with \textit{pre-trained} DistilBERT models, which we fine-tuned for one epoch (i.e. 81.789 optimization steps).\footnote{
    Each model evaluated in this paper was fine-tuned on the pre-trained DistilBERT available via the Hugging Face model hub.}
\end{itemize}

The combination of these choices shrank training time from more than a week to around 24 hours per model. 

With the OCR quality being overall rather questionable, we avoided sentence splitting but chunked the text into units of hundred tokens. We furthermore lower-cased all text but refrained from any additional processing to make downstream comparison easier later on.

The main purpose of our experiments is to estimate how different strategies for appending and masking metadata affect the quality of the language model. 
By analysing in which scenarios MDMA improve the performance of the fine-tuned DistilBERT models, we hope to provide an answer to the following questions:

\begin{itemize}
    \item \textbf{RQ1: Should metadata be added as special tokens?}: As metadata \textit{describes} content, we assess if the year of publication (or other metadata fields) should be prepended as special or normal tokens to the input data. In the former scenario, we added distinct year tokens such ``[1850]'' to the tokenizer and each text fragment. The latter strategy adds that year as normal tokens (i.e. ``1850'').
    Special tokens set the metadata apart from the actual text. Does this make it more difficult for the model to learn a relationship between text and context?
    \item \textbf{RQ2: Should we change the masking probability?}: Does the model perform better if we mask the metadata at higher rates? In the standard scenario, we masked all tokens (including the metadata fields) with the default probability of 0.15. 
    For other models, we use different masking probabilities for text and metadata: we increase the masking probability for the metadata tokens to 0.25 (or 0.75) but continue masking words in the actual text with the default value of 0.15.
    \item \textbf{RQ3: Should we combine the metadata fields?}: When describing a text document, should we add all metadata attributes simultaneously? Or should we just add one feature at a time? Which approach makes the model most sensitive to extra-textual context?
\end{itemize}

To address these questions we trained the following series of models:
\begin{itemize}
    \item \textbf{ERWT series}: The \textbf{E}ncoder \textbf{R}epresentations \textbf{w}ith \textbf{T}ime or \textit{ERWT} models.\footnote{\textit{Erwt} is Dutch for \textit{pea}.} uses the issue's year of publication, e.g. 1810, 1811, as metadata.
    \item \textbf{PEA series}: The \textbf{P}olitical \textbf{E}nhanced represent\textbf{A}tions or \textit{PEA} models\footnote{\textit{Pea} is English for \textit{erwt}.} are trained using the newspaper's declared political orientation---such as liberal or conservative---as recorded in Victorian resources or scholarly summaries \cite{beelen2022bias}.
    \item \textbf{PEA\textsuperscript{TM}}: This model was trained using a combination of metadata fields (year, political leaning and place of publication).
\end{itemize}

Table \ref{table:model_overview} gives an overview of all the DistilBERT models we fine-tuned and evaluated in this paper. 
The \textit{metadata} column records which fields were added to each text segment. \textit{Special token} indicates whether we added the values for these fields as special tokens. With respect to \textit{masking}, we initially used the standard probability of 0.15 over all the tokens in a segment (including the added metadata tokens). This strategy is referred to as NORMAL in table \ref{table:model_overview}. For some models, we experimented with increased metadata masking. For ERWT\_mask\_25, we've hidden the year of publication 25\% of the time and masked the other tokens at the standard rate. 
The last column gives an example of the formatted input data. 
To more accurately assess the impact of adding metadata to the training routine we also created an HMDistilBERT model on the same set of text fragments but without adding the year of publication or other information to the input data. 
This allows us to distinguish between the effects of fine-tuning on historical data and the contributions of MDMA.
\begin{table*}
\begin{tabular}{lllll}
name &     metadata & special token & masking & example input  \\
HMDistilBERT    & NO & NO & NORMAL & Text Fragment... \\
ERWT    & YEAR & NO & NORMAL & 1861 [DATE] ...  \\
ERWT\_st    & YEAR & YES & NORMAL & [1861] [SEP] ... \\
ERWT\_mask\_25    & YEAR & NO & 0.25 & 1861 [DATE] ... \\
ERWT\_mask\_75    & YEAR & NO & 0.75 & 1861 [DATE] ... \\
PEA    & POL & NO & NORMAL & liberal [POL] ... \\
PEA\_st    & POL & YES & NORMAL & [lib] [POL] ... \\
PEA\textsuperscript{TM}   & YEAR, POL, LOC & NO & 0.25 & 1861 [Y] lib [P] lon  [L] ... \\
\end{tabular}
\caption{Overview of the DistilBERT models the models that were fine-tuned for the experiments in this paper.}
\label{table:model_overview}
\end{table*}

Probably undertrained, the ERWT and PEA series are far from perfect. However, we do assume that the \textit{relative differences} in performance are a reliable measure to assess the effects of different preprocessing and masking approaches.\footnote{We trained these models mainly for evaluation purposes but released them also on the HuggingFace Hub: https://huggingface.co/Livingwithmachines.}

\section{Evaluation}

In this section, we investigate in greater detail how different strategies for incorporating metadata play out in downstream applications and tasks: when and how does adding extra-textual knowledge make a difference?

\subsection{Pseudo-Perplexity}

Perplexity is a standard metric to measure the performance of language models.\footnote{
As a reminder: lower perplexity scores indicate better quality.}
Usually, the underlying likelihood is computed by iterating over a sequence and, at each step, calculating the probability of the subsequent token.
However, for MLMs like DistilBERT, we computed the pseudo-perplexity, based on pseudo-log-likelihood (PLL) as described by \cite{salazar2019masked}.
To obtain the PLL score, we mask each token one by one and use the \textit{whole} surrounding context to compute the probability. 
For a document $W$ and model with parameters $\theta$ the PPL is equal to: 

\begin{equation}
PPL(W) := \sum_{t=1}^{|W|} log P_{MLM} (w_{t} | W_{\\t} ; \theta)
\end{equation}
We computed Pseudo Perplexity (PP) for a sequence $W$ of length $N$ as:
\begin{equation}
PP(W) := - \frac{1}{N}\left ( e^{PPL} \right )
\label{equation:pp}
\end{equation}
For our experiments, we made two changes to the PP calculations.
\begin{itemize}
    \item While perplexity is usually calculated over the whole corpus, we wanted to scrutinize in which contexts MDMA enhances the quality of the ERWT models and therefore computed a PP score for each chunk of $N$ tokens separately. We report the average score across all fragments in table \ref{table:perplexity}.\footnote{We kept $N$ constant to either 128 or 64 for each experiment.}
    \item We prepended each segment of $N$ tokens with a year and separator token as required by the model. For the original DistilBERT model we inserted the year and `[SEP]` before the actual text. We then obtain the PP as described in equation \ref{equation:pp} except that we do not mask metadata tokens. For example: to obtain the perplexity for the sequence ``patronage of her majesty the queen'' published in 1841 for the ERWT model, we would, at step one, attach ``1841 [DATE]'' and only start masking at ``patronage'', (i.e. ``1841 [DATE] [MASK] of her Majesty the Queen''). Then move on to the second iteration by hiding the token ``of'' (``1841 [DATE] patronage [MASK] her Majesty the Queen''), etc.
\end{itemize}

\begin{table*}
\begin{tabular}{l | lr | lr}
{length} &      64 &       &       128 &        \\
{model} &      mean &      sd &       mean &       sd \\
DistilBERT            &   354.40   &    376.32  &  229.19 &  294.70 \\
HMDistilBERT          &  32.94 &  64.78 &   25.72 &   45.99 \\
ERWT          &  31.49 &  61.85 &   24.97 &   44.58 \\
ERWT\_st       &  31.69 &  62.42 &   25.03 &   44.74 \\
ERWT\_masked\_25 &  \textbf{30.97} &  61.50 &   \textbf{24.59} &   44.36 \\
ERWT\_masked\_75 &  31.02 &  61.41 &   24.63 &   44.40 \\
PEA &  31.63 &  62.09 &  25.58 &  44.99  \\
PEA\_st &  31.65 &  62.19 &  25.59  & 44.99   \\
\end{tabular}
\caption{Pseudo Perplexity (PP) scores for each DistilBERT model. We compute perplexity 2500 fragments of 64 tokens (left) and 1000 snippets of 128 tokens (right). Lower scores indicate better performance.}
\label{table:perplexity}
\end{table*}

As calculating perplexity was compute-intensive\footnote{Even though not as expensive as normal perplexity, see \cite{salazar2019masked}}, we produced perplexity scores for 1000 fragments of 128 tokens and 2500 chunks of length 64.
We evaluated the models on exactly the same masked text sequences.
Table \ref{table:perplexity} lists the resulting perplexity scores for each model. 
The impact of LM fine-tuning on perplexity scores immediately catches the eye. 
The original DistilBERT model performs very badly on this task, with PP scores almost a magnitude of ten higher than HMDistilBERT.
This begs the question whether fine-tuning really ``historicizes'' language models or adapts them to the peculiarities of textual data with omnipresent OCR errors.
Notwithstanding, the effect on the perplexity scores is substantial.
Interestingly enough HMDistilBERT performs systematically worse than any of the ERWT models, which suggests that inserting metadata in the training routine produces a positive effect, even though it is not comparable to the benefits introduced by fine-tuning on historical data.
The differences are smaller but they are nonetheless significant.\footnote{We compared the scores of HMDistilBERT with the results of the ERWT models using a paired t-test with alternative hypotheses set to ``greater''.}

Looking more closely at how the different time-masking strategies played out, we can observe a slight preference for increased masking and using the standard vocabulary instead of special year tokens.
ERWT\_st has a slightly higher perplexity than the standard ERWT, which is however outperformed, to a larger degree, by ERWT\_masked\_25.
Similarly to \cite{rosin2022time}, the benefits of time-masking don't seem to increase linearly, at least not in this scenario, as ERWT\_masked\_75 scores slightly worse again.

Because we computed the PP for each segment of 64 tokens individually, we can inspect how the score varies as a function of relevant independent variables. 
Using a simple OLS regression model, we modelled the PP as a function of OCR quality\footnote{Please note the OCR quality scores were computed at the article level and thus serve as a rough proxy for the chunk. These scores report the confidence of the OCR software, rather than providing an `objective` extrinsic measure of OCR quality.}, the decade in which the article was published and the political leaning of newspapers.
We created dummy variables for each political label and treated the decade as a categorical variable.\footnote{We used the Python package \textit{statsmodels} for model fitting. The R-style equation looked like: $PP ~ OCR + C(decade) + con + rad + lib$.}
The results are reported in the Appendix, see table \ref{table:ppeffects}.
Obviously, OCR has the largest effect, reminding us of the challenge historical data poses to language models.
For HMDistilBERT we expect the PP score to decrease with 265 when moving the OCR quality from 0 (very bad) to 1 (perfect). 
More importantly, MDMA seem to dampen the impact of OCR. 
The coefficients for the OCR variable shrink when using ERWT models, which suggests that training models with metadata may be more robust to OCR noise, i.e. the quality of the predictions varies less as a function of OCR errors.
In addition, increased masking of metadata seems to produce even more robust models.
However, we have to warn that at this point the gains remain small, and whether they increase when training longer on more data remains to be seen.
\footnote{
We also ran a regression model on the \textit{differences} between HMDistilBERT and each ERWT model, which suggested that these changes in the parameter values are statistically significant.}
Of special interest to us was understanding how perplexity varies in relation to ``representativeness'' of (political) categories in our data: are articles from underrepresented groups, such as the radical press, treated worse by the language model?
We observe that for each political leaning, the coefficients for the ERWT models shrink compared to HMDistilBERT. 
Especially for the radical press, who constituted only a small minority in our dataset, the relative difference is the largest: from 60.49 (for HMDistilBERT) to 55.89 (for ERWT\_masked\_25), a percentage change of around 7.5\%. 
This may indicate that, in general, the perplexity for the radical fragments is higher (i.e. it poses more challenges to the language model compared to liberal papers), but inserting metadata tends to reduce the damage somewhat.
However, more detailed research is needed to understand the actual implications of these findings for historical research.
\footnote{Unfortunately, we did not observe similar consistent effects for the decade variable, this will require further research.}

\subsection{Predicting Metadata with Mask-filling}

ERWT and PEA models are versatile: they listen to the metadata cues, but can also predict aspects of extra-linguistic context. 
In this experiment, we evaluate the latter ability. 
We firstly zoom in on the ERWT series and assess how well these models correctly guess the year of publication given a text fragment.
We do this by masking the metadata token of each fragment and taking the filler with the highest probability as the predicted year of publication. 
Initially, we evaluated our models on a random sample of 10,000 text chunks with a length of 100 tokens. 
All the texts used for evaluation were \textit{not} used for training the language model. 

We created a \textit{random} and \textit{majority} baseline to benchmark ERWT and PEA's mask-filling abilities. 
For the \textit{majority} baseline, we chose the year (or the political label) that appeared most frequently in the training data. 
Given how unbalanced the data are, temporally as well as politically, the most common label is not a trivial baseline to beat.
As a \textit{random} baseline, we sampled randomly over the distribution of years (or political labels) in the training set.
Of course, a model specifically trained for predicting years or political leaning may perform better than masking tokens with ERWT.
However, the point of our experiment is to investigate which preprocessing strategy delivers the strongest model for this task.
By adding the baseline, we only want to demonstrate that the model is learning a non-trivial relation between text and metadata during pre-training.
Later, in section \ref{section_classification}, we scrutinize whether MDMA improves language models for supervised classification. 

\begin{table}
\begin{tabular}{l| r | r| r}
{model} &      random  &    no year & bad ocr   \\
random                  &  19.458 &  &   \textbf{6.678*}/32.343 \\
majority               &  13.929 &  & \textbf{4.488*}/33.452 \\
ERWT           & 8.235   &  8.625  &   7.487 \\
ERWT\_st        & 9.420  &    9.744 &   7.523 \\
ERWT\_masked\_25 &  7.353   &  7.756 &   7.311 \\
ERWT\_masked\_75 & \textbf{6.066}   &   \textbf{6.317} &   \textbf{7.182} \\
\end{tabular}
\caption{Mean absolute error between predicted and actual year using the most probable filler for the masked year token as the predicted year. Results were computed on i.) a random sample (left), ii.) the same sample but with year mentions removed (middle) iii.) articles from before 1830 with low OCR quality (less than 0.5) (right). Scores obtained with a biased baseline are marked with an asterisk.}
\label{table:yearprediction}
\end{table}

Table \ref{table:yearprediction} reports the results for masking the time metadata token.
We calculated the mean absolute error between the predicted and true year of publication.\footnote{We did look into the top $n$ predictions for each masked metadata token, but this didn't change our findings and interpretation, and we, therefore, decided to remove these numbers from the paper to avoid the tables becoming too unwieldy.}
All ERWT models yield substantially better predictions than the baselines.
When given a snippet of 100 words, the mean distance between the predicted and actual timestamp varies between six or eight years, which is vastly better than the baselines, which meander between 14 and 20 years.
Put differently, comparing the worst ERWT model with the best baseline, the error term increases by roughly 70\% (from ca. 8.2 to 13.9 years). 
This clearly shows that by systematically appending temporal metadata ERWT learns to situate documents in time.  
With respect to tokenization strategy, the results indicate that adding distinctive year tokens to the vocabulary slightly hurts the performance. 
The simpler preprocessing approach is approximately one year closer to the actual year, which amounts to a small but significant difference ( using a two-sided t-test). 
More importantly, table \ref{table:yearprediction} suggests that for the task of mask-filling, the more heavily masked models deliver better results. 
The average error drops to just above 6 years for ERWT\_masked\_75, which is well above its nearest competitor ERWT\_masked\_25.
Of course, a sceptical reader may rightfully doubt that ERWT really learns to relate texts to time. 
Maybe the model relies on  explicit markers, such as the year mentioned in a text, to predict the date of publication.
To evaluate this, we reran the above experiments but stripped all references to years from the text.\footnote{We used the following regular expression $r'[789][0-9]{2}'$ and substituted all matched patterns with white space.}
As expected, ERWT's performance drops, but only minimally so. 
Across all cases, the model trained without special year tokens obtains the best scores.
The sample is referred to as ``no year'' in table \ref{table:yearprediction}.

To challenge our undertrained model even more, we evaluated its performance on articles from before 1830 with poor OCR quality, also removing year references (the ``bad ocr'' sample).
The scores obtained from this sample are not comparable with the \textit{random} set.
We gave the baselines access to information the language model does not have: the \textit{biased majority} baseline takes the most frequent year appearing in the training set \textit{before} 1830. The \textit{biased random} approach similarly samples year labels from examples before 1830.
In other words: these baselines \textit{know} that all snippets date from before 1830, information ERWT does not have access to, as it was trained on the whole 1800-1880 period.
Scores for biased baselines are marked with an asterisk.
We also report the results for the non-biased baselines which sample over the complete year distribution.
Luckily, the quality does not deteriorate radically. 
Even in this situation, the ERWT models perform reasonably well: the scores are a bit worse than those of the biased baselines, but well below the unbiased ones.
In any case, these results show that avoiding special year tokens and using the standard vocabulary yields the best results.\footnote{Relying on the tokenizer's standard vocabulary has one slight drawback: the model is more likely to return tokens that are not years.}

By training the PEA models, we also experimented with inserting political information into the language model.
We followed the same training strategy as above, prepending the political orientation of a newspaper either as a standard or a special token.
Based on the same sample of 10,000 snippets, we masked the token indexing the political leaning and used the filler with the highest probability as the prediction.  
Again, we used \textit{majority} and \textit{random} as baselines.
Given that HMD is a very unbalanced dataset (the vast majority of the articles are from liberal papers), we used macro f1, precision and recall as metrics for evaluation.

\begin{center}
\begin{table*}
\begin{tabular}{l|lrr| lrr}
{sample} &  random &   &   & political &   &   \\
{model} &  macro f1 &  precision &  recall & macro f1 &  precision &  recall \\
random       &     0.198 &            0.199 &         0.197  &     0.204 &            0.203 &         0.210 \\
majority     &     0.171 &            0.150 &         0.200 &     0.172 &            0.151 &         0.200 \\
PEA    &     0.474 &            0.665 &         0.413 &     0.469 &            0.668 &         0.412 \\
PEA\_st &     0.461 &            0.655 &         0.397 &     0.467 &            0.709 &         0.397 \\
PEA\textsuperscript{TM}   &     \textbf{0.475} &            0.664 &         0.415 &     \textbf{0.470} &            0.713 &         0.402 \\
\end{tabular}
\caption{Prediction of political leaning based on the most probable filler for the masked metadata token. Results were computed on i.) a random sample (left), ii. a sample that contained politically charged keywords (right). }
\label{table:polpredict}
\end{table*}
\end{center}

In terms of their f1 scores, the PEA models did not perform very well, but they were still substantially better than the baselines.
Similarly to the ERWT series, the masking strategy that relied on special tokens to index metadata performed somewhat worse.
We anticipated that the PEA\textsuperscript{TM} would outperform its more simplistic peers, but to this was hardly the case.
By observing multiple metadata fields during pre-training, we expected the model to better handle the confounding of the politics, place and time.
Put differently, we thought combined metadata might help PEA\_\textsuperscript{TM} to establish which aspects of language use are symptomatic of the Metropolitan press and which reflect a paper's liberal orientation.
However, the gains from PEA\textsuperscript{TM} were only minimal, unfortunately. 

The absence of partisan signals in the data may also explain the low scores.
To evaluate this, we created a sample that contained at least one ideologically charged ``political'' keyword.\footnote{We used snippets that contained the following terms: liberal(s), conservative(s) and tory/ies. From previous experiments we knew that they are strongly related to the political leaning of the newspaper.}
Also these results---again somewhat to our surprise---did not change to a great extent.
PEA still beats both the random and majority baselines but it does not obtain higher f1 scores compared to the random sample.
A consistent finding across all these experiments was that adding metadata as special tokens tends to hurt performance. 
Using the tokenizer's standard vocabulary is the better option.

\subsection{Supervised Experiments: Classification and Regression} \label{section_classification}

In this section, we evaluate how ERWT and PEA models perform on supervised classification tasks with historical data.
More precisely we gauge the extent to which our models can predict political leaning, date, and animacy.
For each experiment, we load the pre-trained language models and change the classification head to perform either multi-class classification (political leaning), binary classification (animacy) or regression (time).

To detect partisanship in newspaper articles, we used excerpts of hundred tokens that were not part of the language modelling task and trained a classifier to predict the political leaning of the newspaper in which the snippet appeared.
Again, we care more about relative than absolute performance on this task. 
We firstly trained models on a random sample, but because (as observed above) 
many fragments lacked partisan signals, we also evaluated the classifiers on fragments that contained political keywords, such as ``liberal'' and ``conservative''.
Both the \textit{random} and the \textit{political} sample contain 15.000 fragments. 
10200 were used for training, 1800 for validation and the remaining 3000 for testing.
Each text snippet was appended with year metadata and special tokens as required by the model.
We used the same number of epochs (10) and learning rate (0.0002) for each pre-trained model, to ensure the comparison is fair.

\begin{table*}
\begin{tabular}{l| rr|rr}
{sample} &  random &     &  political &      \\
{model} &  f1 macro &  accuracy &  f1 macro &    accuracy \\
DistilBERT             &     0.631  &     0.800       &     0.639 &         0.822 \\
HMDistilBERT       &     0.663 &        0.828   &     0.694 &     0.843 \\
ERWT          &     0.667 &        0.830   &     0.692 &         0.846 \\
ERWT\_st        &     0.670 &       0.820   &     0.691 &         0.846 \\
ERWT\_masked\_25 &     0.671 &        0.831 &     0.690 &         0.847 \\
ERWT\_masked\_75 &     \textbf{0.672} &         0.831  &     0.688 &         0.846 \\
PEA              &     0.662 &        0.825     &     0.683 &        0.845 \\
PEA\_st           &     0.668 &       0.827    &     0.682 &        0.844 \\
PEA\textsuperscript{TM} & \textbf{0.672} &        \textbf{0.833}  &     \textbf{0.691} &            \textbf{0.850} \\
\end{tabular}
\caption{Performance of classifier trained on ca. 10,000 text snippets of hundred tokens and political labels. We trained the classifier both on a random sample (left) and a sample in which each snippet contained a political keyword (right).}
\label{table:polclassification}
\end{table*}

As can be surmised from table \ref{table:polclassification}, we remain uncertain about the merits of MDMA for downstream classification tasks.
Pre-training on historical data does seem to improve accuracy in general, however, the differences between HMDistilBERT and the other models appear to be small and not consistently in the favour of the ERWT/PEA series (especially for the random sample).
From these preliminary results, we gather that models which apply intenser time-masking may perform better, with ERWT\_maked\_25 obtaining the highest f1 and accuracy scores for this type of model.
The models trained by only inserting political labels perform worse than those that used time-masking——again somewhat to our surprise. 
At the moment we don't have a clear explanation.
An issue here could be that the hyperparameters, such as the learning rate, destroy valuable knowledge present in the fine-tuned models.
The strongest performer, albeit by a small margin, is PEA\_\textsuperscript{TM}.
Contrary to our earlier findings, combined metadata may make language models better for sequence classification.

To assess how models trained with MDMA, perform on other genres---outside of the newspaper universe---we evaluated them on \textit{atypical animacy} dataset \cite{Tolfo2020LivingMA, ardanuy2020living}. This corpus contains annotations on the animacy of machines, i.e. we asked annotators if the machine in a given sentence was portrayed as ``alive'' or not? 
The data is derived from a collection of 19th Century books.\footnote{https://www.bl.uk/collection-guides/digitised-printed-books} 
Given that this dataset is considerably smaller, we reduced the number of epochs (to three) as well as the learning rate (to 0.0001).

\begin{table*}
\begin{tabular}{l | rr}
{model} &  f1 binary &   accuracy \\
DistilBERT             &      0.752 &       0.826 \\
HMDistilBERT         &      0.741 &       0.831 \\
ERWT           &      0.769 &       0.848 \\
ERWT\_st        &      0.762 &        0.831 \\
ERWT \_masked\_25 &     \textbf{0.777} &        \textbf{0.848} \\
ERWT \_masked\_75 &      0.756 &        0.837 \\
PEA              &      0.752 &       0.837 \\
PEA\_st           &      0.763 &       0.843 \\
PEA\_\textsuperscript{TM} &      \textbf{0.777} &        \textbf{0.848} \\
\end{tabular}
\caption{Results for training a sequence classifier on the Atypical Animacy dataset.}
\label{table:animacy}
\end{table*}


Table \ref{table:animacy} largely paints a similar picture as we observed in the previous classification experiment.
There seems something to gain from pre-training on historical texts with added metadata, but not that much.
ERWT\_masked\_25 and PEA\_\textsuperscript{TM} appear as the strongest competitors among the fine-tuned models, but not miles ahead even compared to the standard DistilBERT model.
In general, the improvement in accuracy scores meanders around 2\% between the off-the-shelf model and the best ERWT/PEA variant. 

Lastly, we experimented with year regression (see results in the Appendix) but our findings tied in with the previous two experiments. Masking the year field, yielded actually better scores than training a supervised classifier on top of the ERWT pre-trained model. 
\footnote{As a disclaimer, we need to add that in contrast to the previous evaluation tasks, supervised results are harder to reproduce and can vary just by changing the random seed. The supervised models are moreover very sensitive to hyperparameters and require further study to become more robust.}

\section{Conclusions and Future Work}

So, is MDMA all you need for training better language models? 
The findings in this paper demonstrated that adding metadata can make language models more sensitive to aspects of extra-textual context.
and they appeared to be more robust to bias in data and OCR noise (as suggested by the analysis of pseudo-perplexity).
Also, MDMA adds new functionality to MLMs.
ERWT, for example, could accurately situate fragments of newspaper texts in time (even though PEA models struggled with detecting partisanship).
Experiments with supervised machine learning suggested that pre-training with metadata likely leads to better accuracy scores, but the evidence remains mixed and somewhat fragile.

In general, this paper allows us to propose the following guidelines for metadata masking:
\begin{itemize}
    \item \textbf{RQ1:} adding metadata as special tokens tends to diminish performance on our evaluation tasks.
    \item \textbf{RQ2:} masking metadata at higher rates (i.e. 25\% of the time) generally delivered the best results. 
    \item \textbf{RQ3:} combining metadata may provide more robust models for sequence classification.
\end{itemize}

Since adding metadata to the training process is ``cheap'' in comparison to other modifications, even modest improvements to the performance of the model may warrant the use of this modified approach.

In future work, we will train models on a larger and more diverse newspaper dataset (in terms of their political and geographical composition,) but also include other collections such as books. 
We started to experiment with approaches that require changes to the objective function (not just the input data) as well.
Lastly, we aim to extend our evaluation to different tasks that use historical (newspaper) data, such as Named Entity Recognition or emotion classification.

\section{Appendix}

\begin{center}
\begin{table*}
\begin{tabular}{llrr}
 parameter              &         model       &     coefficient &  p-value \\
ocr & HMDist & -264.930 &   0.000 \\
               & ERWT & -253.228 &   0.000 \\
               & ERWT\_ts & -255.035 &   0.000 \\
               & ERWT\_masked\_25 & -250.516 &   0.000 \\
               & ERWT\_masked\_75 & -250.432 &   0.000 \\
pol\_rad & HMDist &   60.481 &   0.000 \\
               & ERWT &   56.519 &   0.000 \\
               & ERWT\_ts &   57.072 &   0.000 \\
               & ERWT\_masked\_25 &   55.877 &   0.000 \\
               & ERWT\_masked\_75 &   55.967 &   0.000 \\
pol\_con & HMDist &   37.560 &   0.000 \\
               & ERWT &   36.074 &   0.000 \\
               & ERWT\_ts &   36.397 &   0.000 \\
               & ERWT\_masked\_25 &   35.526 &   0.000 \\
               & ERWT\_masked\_75 &   35.573 &   0.000 \\
pol\_lib & HMDist &   36.718 &   0.000 \\
               & ERWT &   35.115 &   0.000 \\
               & ERWT\_ts &   35.461 &   0.000 \\
               & ERWT\_masked\_25 &   34.641 &   0.000 \\
               & ERWT\_masked\_75 &   34.694 &   0.000 \\
dec\_1820 & HMDist &   -8.950 &   0.125 \\
               & ERWT &   -6.989 &   0.210 \\
               & ERWT\_ts &   -7.462 &   0.186 \\
               & ERWT\_masked\_25 &   -6.462 &   0.245 \\
               & ERWT\_masked\_75 &   -6.602 &   0.234 \\
dec\_1860 & HMDist &   10.780 &   0.062 \\
               & ERWT &   12.088 &   0.029 \\
               & ERWT\_ts &   11.750 &   0.035 \\
               & ERWT\_masked\_25 &   12.291 &   0.026 \\
               & ERWT\_masked\_75 &   11.990 &   0.029 \\

\end{tabular}
\caption{shows the coefficients and p-values obtained by running an OLS regression model with pseudo-perplexity as the dependent variable and decade, political leaning and OCR quality as predictors.}
\label{table:ppeffects}
\end{table*}
\end{center}

\begin{center}
\begin{table*}
\begin{tabular}{lrrrr}
{}             &     loss &     mean absolute error &      mean squared error &  accuracy \\
DisilBERT &  354.606 &  15.148 &  354.606 &     0.022 \\
HMDistilBERT &  354.235 &  15.323 &  354.235 &     0.022 \\
ERWT &  313.965 &  14.556 &  313.965 &     0.018 \\
 ERWT\_masked\_25 &  323.114 &  14.469 &  323.114 &     0.020 \\
ERWT\_masked\_75 &  338.570 &  15.287 &  338.570 &     0.014 \\

\end{tabular}
\caption{shows the error and accuracy for predicting the year of publication given a text fragment}
\label{table:timereg}
\end{table*}
\end{center}


\bibliography{bibliography}
\bibliographystyle{plainnat}


\end{document}